\documentclass[11pt,a4paper]{article}
\usepackage[hyperref]{emnlp2020}
\usepackage{times}
\usepackage{latexsym}

\usepackage{microtype}
\usepackage{rotating}
\usepackage{lipsum}
\usepackage{xcolor}
\usepackage{enumitem}

\aclfinalcopy 
\def\aclpaperid{3302} 

\title{Visually-Grounded Planning without Vision: \\ Language Models Infer Detailed Plans from High-level Instructions}

\author{Peter A. Jansen \\
  School of Information, University of Arizona, Tucson, AZ\\
  \texttt{pajansen@email.arizona.edu} }

\date{}

\begin{document}
\maketitle
\begin{abstract}
The recently proposed \textsc{ALFRED} challenge task aims for a virtual robotic agent to complete complex multi-step everyday tasks in a virtual home environment from high-level natural language directives, such as ``put a hot piece of bread on a plate''.  Currently, the best-performing models are able to complete less than 5\% of these tasks successfully.  In this work we focus on modeling the translation problem of converting natural language directives into detailed multi-step sequences of actions that accomplish those goals in the virtual environment.  We empirically demonstrate that it is possible to generate gold multi-step plans from language directives alone without any visual input in 26\% of unseen cases.  When a small amount of visual information is incorporated, namely the starting location in the virtual environment, our best-performing GPT-2 model successfully generates gold command sequences in 58\% of cases.  Our results suggest that contextualized language models may provide strong visual semantic planning modules for grounded virtual agents.
\end{abstract}

\section{Introduction}



Simulated virtual environments with steadily increasing fidelity are allowing virtual agents to learn to perform high-level tasks that couple language understanding, visual planning, and embodied reasoning through sensorimotor grounded representations \cite{gordon2018iqa,puig2018virtualhome,Wijmans_2019_CVPR}. 
The \textsc{ALFRED} challenge task recently proposed by Shridhar et al. \shortcite{shridhar2019alfred} requires a virtual robotic agent to complete everyday tasks (such as \textit{``put cold apple slices on the table''}) in one of 120 interactive virtual home environments by generating and executing complex visually-grounded semantic plans that involve movable objects, irreversible state changes, and an egocentric viewpoint.  Integrating natural language task directives with one of the most complex interactive virtual agent environments to date is challenging, with the current best performing systems successfully completing less than 5\% of \textsc{ALFRED} tasks in unseen environments\footnote{\url{https://leaderboard.allenai.org/alfred/}}, while common baseline models generally complete less than 1\% of tasks successfully. 

\begin{figure}
\centering
\includegraphics[scale=1.0]{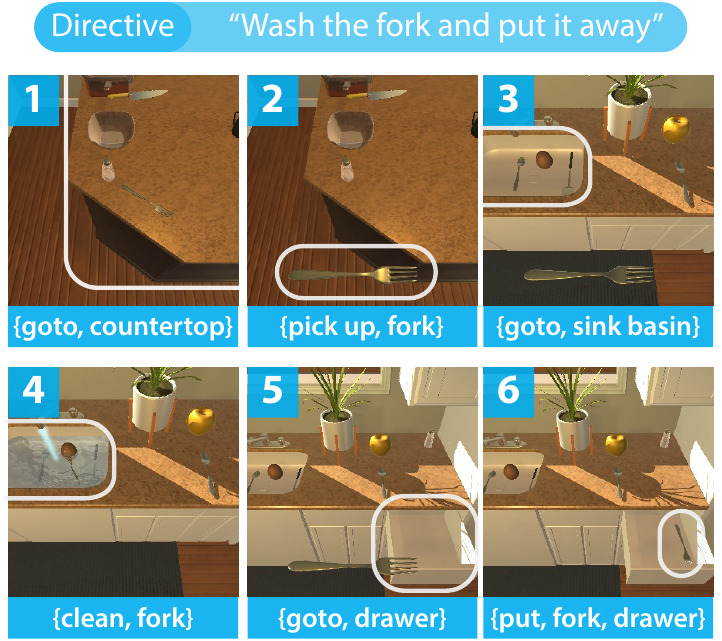}
\caption{\label{tab:example}\footnotesize An example of the \textsc{ALFRED} grounded language task.  In this work, we focus on visual semantic planning -- from the textual directive alone \textit{(top)}, our model predicts a visual semantic plan of \textit{\{command, argument\}} tuples \textit{(captions)} that matches the gold plan without requiring visual input \textit{(images)}.}
\vspace{-4mm}
\end{figure}

In this work we explore the visual semantic planning task in \textsc{ALFRED}, where the high-level natural language task directive is converted into a detailed sequence of actions in the \textsc{AI2-THOR 2.0} virtual environment \cite{kolve2017ai2} that will accomplish that goal (see Figure~\ref{tab:example}). In contrast to previous approaches to visual semantic planning \cite[e.g.][]{zhu2017visual,fried2018speaker,fang2019scene}, we explore the performance limits of this task solely using goals expressed in natural language as input -- that is, \textit{without} visual input from the virtual environment.  The contributions of this work are:
\begin{enumerate}
\item We model visual semantic planning as a sequence-to-sequence translation problem, and demonstrate that our best-performing GPT-2 model can translate between natural language directives and sequences of gold visual semantic plans in 26\% of cases without visual input. 
\item We show that when a small amount of visual input is available -- namely, the starting location in the virtual environment -- our best model can successfully predict 58\% of unseen visual semantic plans.
\item Our detailed error analysis suggests that repairing predicted plans with correct locations and fixing artifacts in the \textsc{ALFRED} dataset could substantially increase performance of this and future models. 
\end{enumerate}



\section{Related Work}

Models for completing multi-modal tasks can achieve surprising performance using information from only a single modality.  The Room-to-Room (R2R) visual language navigation task \cite{anderson2018vision} requires agents to traverse a discrete scene graph and arrive at a destination described using natural language.  In ablation studies, Thomason et al.~\shortcite{thomason-etal-2019-shifting} found that models using input from a single modality (either vision or language) often performed nearly as good as or better than their multi-modal counterparts on R2R and other visual QA tasks.  Similarly, Hu et al.~\shortcite{hu-etal-2019-looking} found that two state-of-the-art multi-modal agents performed significantly worse on R2R when using both linguistic and visual input instead of a single modality, while also showing that performance can improve by combining separate-modality models into mixture-of-expert ensembles. 

Where R2R requires traversing a static scene graph using locomotive actions, \textsc{ALFRED} is a dynamic environment requiring object interaction for task completion, and has a substantially richer action sequence space that includes 8 high-level actions.  This work extends these past comparisons of unimodal vs. multimodel performance by demonstrating that strong performance on visual semantic planning is possible in a vastly more complex virtual environment using language input alone, through the use of generative language models.   





%
%
\begin{table*}[!t]
\centering
\footnotesize
\begin{tabular}{llccccccccccc}
	& 		& &	\multicolumn{3}{c}{Triple Components}	& & 	Full	& & \multicolumn{2}{c}{Entire Visual Semantic Plans}	\\
	& Model & &	Command	& Arg1		& Arg2		&&	Triples		& &	Full Sequence	&	Full Minus First
 \\[1mm] \hline 
 \multicolumn{7}{l}{\textit{Strict Scoring}}	\\
	& RNN	& & 	89.6\%	&	64.8\%	&	58.4\%	& &	60.2\%	& &	17.1\%	&	43.6\%	 \\[1mm]
	& GPT-2	& & 	90.8\%	&	69.9\%	&	63.8\%	& &	65.8\%	& &	22.2\%	&	53.4\%	 \\[1mm] 
	 
 \multicolumn{7}{l}{\textit{Permissive Scoring}}	\\
	& RNN	& & 	89.6\%	&	70.6\%	&	61.4\%	& &	65.9\%	& &	23.6\%	&	51.6\%	 \\[1mm]
	& GPT-2	& & 	90.8\%	&	73.8\%	&	65.1\%	& &	69.4\%	& &	26.1\%	&	58.2\%	 \\[1mm] \hline
~\\[1pt]
\end{tabular}
\vspace{-4mm}
\caption{\label{tab:performance} \footnotesize Average prediction accuracy on the unseen test set broken down by triple components, full triples, and full visual semantic plans. \textit{Full Sequence} accuracy represents the proportion of predicted visual semantic plans that perfectly match gold plans.  \textit{Full Minus First} represents the same, but omitting the first tuple, typically a \textit{\{goto, location\}} that moves the agent to the starting location in the virtual environment (see description in text).}
\end{table*}

%
%
\begin{table}[!t]
\centering
\footnotesize
\vspace{3mm}
\begin{tabular}{lccccccccc}
Model &
\begin{rotate}{60} Goto \end{rotate} &
\begin{rotate}{60} Pickup \end{rotate} &
\begin{rotate}{60} Put \end{rotate} &
\begin{rotate}{60} Cool \end{rotate} &
\begin{rotate}{60} Heat \end{rotate} &
\begin{rotate}{60} Clean \end{rotate} &
\begin{rotate}{60} Slice \end{rotate} &
\begin{rotate}{60} Toggle \end{rotate} &
\begin{rotate}{60} Avg. \end{rotate} \\[1mm] \hline 
\vspace{-2mm}
~\\[1pt]
RNN		& 	59			&	81			&	60			&	\textbf{77}	&	69			&	\textbf{83}	&	67			&	91			& 	66 			\\[1mm]
GPT-2	& 	\textbf{63}	&	\textbf{84}	&	\textbf{66}	&	72			&	\textbf{77}	&	82			&	\textbf{70}	&	\textbf{94}	& 	\textbf{69} \\[1mm]  \hline
\end{tabular}
\caption{\label{tab:performancebycommand} \footnotesize Average triple prediction accuracy on the test set broken down into each of the 8 possible \textsc{ALFRED} commands. Values represent percentages. \textit{Goto} has an $N$ of 24k, \textit{Pick up} an $N$ of 11k, and \textit{Put} an $N$ of 10k.  All other commands occur approximately 1000 times in the test dataset.}
\end{table}

\section{Models and Embeddings}
We approach the task of converting a natural language directive into a \textit{visual semantic plan} -- a series of commands that achieve that directive in a virtual environment -- as a purely textual sequence-to-sequence translation problem, similar to conversion from Text-to-SQL \cite[e.g.][]{yu2018spider,guo2019towards}.  Here we examine two embedding methods that encode language directives and decode command sequences. 

{\flushleft\textbf{RNN:}} A baseline encoder-decoder network for sequence-to-sequence translation tasks \cite[e.g.][]{bahdanau2015neural}, implemented using recurrent neural networks (RNNs).  One RNN serves as an encoder for the input sequence, here the tokens representing the natural language directive.  A decoder RNN network with attention uses the context vector of the encoder network to translate into output sequences of command triples representing the visual semantic plan.  Both encoder and decoder networks are pre-initialized with 300-dimensional GLoVE embeddings \cite{pennington2014glove}.

{\flushleft\textbf{GPT-2:}} 
The OpenAI GPT-2 transformer model \cite{radford2019language}, used in a text generation capacity.  
We fine-tune the model on sequences of natural languge directives paired with gold command sequences separated by delimiters (i.e. \textit{``$<$Directive$>$ [SEP] $<$CommandTuple$_{1}$$>$ [CSEP] $<$CommandTuple$_{2}$$>$ [CSEP] ... [CSEP] $<$CommandTuple$_{N}$$>$ [EOS]''}). 
During evaluation we provide the prompt \textit{``$<$Directive$>$ [SEP]''}, and the model generates a command sequence until producing the end-of-sequence (EOS) marker.  We make use of nucleus sampling \cite{holtzman2020curious} to select only tokens from the set of most likely tokens during generation, with $p=0.9$, but do not make use of top-K filtering \cite{fan2018hierarchical} or penalize repetitive n-grams, which are commonly used in text generation tasks, 
but are inappropriate here for converting to the often repetitive (at the scale of bigrams) command sequences.  For tractability we make use of the GPT-2 Medium pre-trained model, which contains 24 layers, 16 attention heads, and 325M parameters.  During evaluation, task directives are sorted into same-length batches to prevent generation artifacts from padding, and maintain high generation quality.\footnote{Negative results not reported for space: We hypothesized that separating visual semantic plans into variablized action-sequence templates and variable-value assignments represented as separate decoders would help models learn to separate the general formula of action sequences with specific instances of objects in action sequences, which has been shown to help in Text-to-SQL translation \cite{guo2019towards}.  Pilot experiments with both RNNs and transformer models yielded slightly lower results than vanilla models.  Language modeling: In addition to GPT-2 we also piloted XLNET, but perplexity remained high even after significant fine-tuning. }


%
%
\begin{table*}[t]
\centering
\footnotesize
\begin{tabular}{cll}
	Prop.	&	Error Class Description	&	Example Errors		 \\[1mm] \hline 
\vspace{-2mm}
~\\[1pt]

\multicolumn{2}{l}{\textit{Incorrect Arguments}}	&	\textit{Predicted wrong location:}\\
	45\%	&	Predicted wrong location			&	(G) ... slice lettuce, put knife on \textbf{countertop}, put lettuce in fridge, ... \\
	4\%		&	Predicted wrong object				& 	(P) ... slice lettuce, put knife in \textbf{microwave}, put lettuce in fridge, ... \\
~\\
\multicolumn{2}{l}{\textit{Incorrect Triples}}					&	\textit{Predicted extra (not harmful) action$^{\dagger}$, and introduced offset error$^{\ddagger}$}\\
	22\%	&	Offset due to extra/missing actions				&	Instructions: Put a mug with a spoon in the sink. \\
	22\%	&	Predicted extra (incorrect) actions				&	(G) ... pick up mug, \textbf{put mug in sink basin$^{\ddagger}$}\\
	12\%	&	Predicted missed actions						&	(P) ... pick up  mug, \textbf{go to sink basin$^{\dagger}$,	put mug in sink basin$^{\ddagger}$}\\
	7\%		&	Predicted extra (not harmful) actions			&	\\
	5\%		&	Order of actions swapped						&	\\
	
~\\
\multicolumn{2}{l}{\textit{Instruction Errors}}					&	\textit{Gold Instructions Incomplete:}\\
	17\%	&	Gold Instructions Incorrect						&	Instructions: Put a heated mug in the microwave.\\
	13\%	&	Gold Instructions Incomplete					&	(G) ... go to microwave, heat mug, \textbf{go to cabinet, put mug in cabinet} \\
\vspace{-2.5mm}	\\ \hline
\end{tabular}
\caption{\label{tab:erroranalysis} \footnotesize \textit{(left)} Common classes of prediction errors in the GPT-2 model, and their proportions in 100 predictions from the development set.  \textit{(right}) Example errors, where (G) and (P) represent subsets of gold and predicted visual semantic plans, respectively.}
\vspace{-3.5mm}
\end{table*}

\section{Experiments}

{\flushleft\textbf{Dataset:}} The \textsc{ALFRED} dataset contains 6,574 gold command sequences representing visual semantic plans, 
each paired with 3 natural language directives describing the goal of those command sequences (e.g. `\textit{`put a cold slice of lettuce on the table''}) 
authored by mechanical turkers.  High-level command sequences range from 3 to 20 commands (average 7.5), and are divided into 7 high-level categories (such as \textit{examine object in light, pick two objects then place,} and \textit{pick then cool then place}). Commands are represented as triples that pair one of 8 actions (\textit{goto, pickup, put, cool, heat, clean, slice, and toggle}) with up to two arguments, typically the object of the action (such as ``slicing \textit{lettuce}'') and an optional receptacle (such as ``putting a spoon in a \textit{mug}''). Arguments can reference 58 possible objects (e.g. \textit{butter knife, chair,} or \textit{apple}) and 26 receptacles (e.g. \textit{fridge, microwave,} or \textit{bowl}).  
To prevent knowledge of the small unseen test set for the full task, 
here we redivide the large training set into three smaller train, development, and test sets of 7,793, 5,661, and 7,571 gold-directive/command-sequence pairs, respectively. 


{\flushleft\textbf{Processing Pipeline:}} Command sequences are read in as sequences of \textit{$\{$command, arg1, arg2$\}$} triples, converted into natural language using completion heuristics (e.g. \textit{``$\{$put, spoon, mug$\}$''} $\rightarrow$ \textit{``put the spoon in the mug''}, and augmented with argument delimiters to aid parsing (e.g. \textit{``put $<$arg1$>$ the spoon $<$arg2$>$ in the mug''}). Input directives are tokenized, but receive no other preprocessing.  Generated strings from all models are post-processed for common errors in sequence-to-sequence models, including token doubling, completing missing bigrams (e.g. \textit{``pick $<$arg1$>$''} $\rightarrow$ \textit{``pick up $<$arg1$>$''}), and heuristics for adding missing argument tags.  Post-processed output sequences are then parsed and converted back into \textit{$\{$command, arg1, arg2$\}$} tuples for evaluation.


{\flushleft\textbf{Evaluation Metrics:}} Performance in translating between natural language directives and sequences of command triples is evaluated in terms of accuracy at the command-element (\textit{command, argument1, argument2}), triple, and full-sequence level.  Because our generation includes only textual input and no visual input for a given virtual environment, commands may be generated that reference objects that do not exist in a scene (such as generating an action to toggle a \textit{``lamp''} to examine an object, when the environment specifically contains a \textit{``desk lamp''}).  As such we include two scoring metrics: a \textit{strict} metric that requires exact matching of each token in an argument to be counted as correct, and a \textit{permissive} metric that requires matching only a single token within an argument to be correct.


\vspace{2mm}
\begin{tabular}{lc}
\hline
\textit{Strict Scoring}		&  butter knife $\ne$ knife	\\[1mm]
\textit{Permissive Scoring} 	&	desk lamp $=$ lamp	\\
\hline			
\end{tabular}
~\\

All accuracy scoring is binary. Triples receive a score of one if all elements in a given gold and predicted triple are identical, and zero otherwise.  Full-sequence scoring directly compares \textit{$<$CommandTuple$_{i}$$>$} 
for each $i$ in the gold and predicted sequences, and receives a score of one only if all triples are identical and in identical locations $i$, and zero otherwise.\footnote{Tuning and Computational Resources: RNN models required approximately 100k epochs of training to reach convergence over 12 hours, requiring 8GB of GPU RAM.  GPT-2 models asymptoted performance at 25 epochs, requiring 6 hours of training and 16GB of GPU RAM.  All experiments were conducted using an NVIDIA Titan RTX. }


\subsection{Results}

Performance of the embedding models is reported in Table~\ref{tab:performance}, broken down by triple components, full triples, and full sequences. Both models achieve approximately 90\% accuracy in predicting the correct commands, in the correct location $i$ in the sequence.  Arguments are predicted less accurately, with the RNN model predicting 65\% and 58\% of first and second arguments correctly, respectively.  The GPT-2 model increases performance on argument prediction by approximately +5\%, reaching 70\% and 64\% under strict match scoring.  Permissive scoring, allowing for partial matches between arguments (e.g. ``lamp'' and ``desk lamp'' are considered equivalent) further increases argument scoring to approximately 74\% and 65\% in the best model.  Scoring by complete triples in the correct location $i$ shows a similar pattern of performance, with the best-scoring GPT-2 model achieving 66\% accuracy using strict scoring, and 69\% under permissive scoring, with triple accuracy broken down by command shown in Table~\ref{tab:performancebycommand}.

Fully-correct predicted sequences of commands that perfectly match gold visual semantic plans using only the text directives as input, -- i.e. \textit{without} visual input from the virtual environment -- occur in 17\% of unseen test cases with the RNN model, and 22\% of cases with the GPT-2 model, highlighting how detailed and accurate visual plans can be constructed from text input alone in a large subset of cases.  In analyzing the visual semantic plans, the first command is typically to move the virtual agent to a starting location that contains the first object it must interact with (for example, moving to \textit{the countertop}, where a potato is resting in the initialized virtual environment, to begin a directive about \textit{slicing, washing, and heating a potato slice}).  If we supply the model with this single piece of visual information from the environment, full-sequence prediction accuracy for all models more than doubles, increasing to 53\% in the strict condition, and 58\% with permissive scoring, for the best-performing GPT-2 model.


\subsection{Error Analysis}

Table~\ref{tab:erroranalysis} shows an analysis of common categories of errors in 100 directive/visual semantic plan pairs randomly drawn from the development set that were not answered correctly by the best-performing GPT-2 model that includes the starting location for the first step.  As expected, a primary source of error is the lack of visual input in generating the visual plans, with the most common error, \textit{predicting the wrong location in an argument}, occuring in 45\% of errors.\footnote{An unexpected source of error is that our GPT-2 planner frequently prefers to store used cutlery in either the fridge or microwave -- creating a moderate fire hazard.  Interestingly, this behavior appears learned from the training data, which frequently stores cutlery in unusual locations.  Disagreements on discarded cutlery locations occurred in 15\% of all errors.}  Conversely, predicting the wrong object to interact with occurred in only 4\% of errors, as this information is often implicitly or explicitly supplied in the text directive.  This suggests augmenting the model with object locations from the environment could mend prediction errors in nearly half of all errorful plans. 

The GPT-2 model predicted additional (incorrect) actions in 22\% of errorful predictions, while missing key actions in 12\% of errors, causing offset errors in sequence matching that reduced overall performance in nearly a quarter of cases.  In a small number of cases, the model predicted extra actions that were not harmful to completing the goal, or switched the order of sets of actions that could be completed independently (such as picking up and moving two different objects to a single location).  In both cases the virtual agent would likely have been successful in completing the directive if following these plans. 

A final significant source of error includes inconsistencies in the crowdsourced text directives or gold visual semantic plans themselves. In 17\% of errors, the gold task directive had a mismatch with the objects referenced in the gold commands (e.g. the directive referenced a \textit{watering can}, where the gold annotation references a \textit{tea pot}), and automated scoring marked the predicted sequence as incorrect.  Similarly, in 13\% of cases, the task directive failed to mention one or more subtasks (e.g. the directive is \textit{``turn on a light''}, but the gold command sequence also includes first retrieving a specific object to examine in the light).  This suggests that nearly one-third of errors may be due to issues in the evaluation data, and that overall visual semantic plan generation performance may be significantly higher.

\begin{figure}
\centering
\includegraphics[scale=0.30]{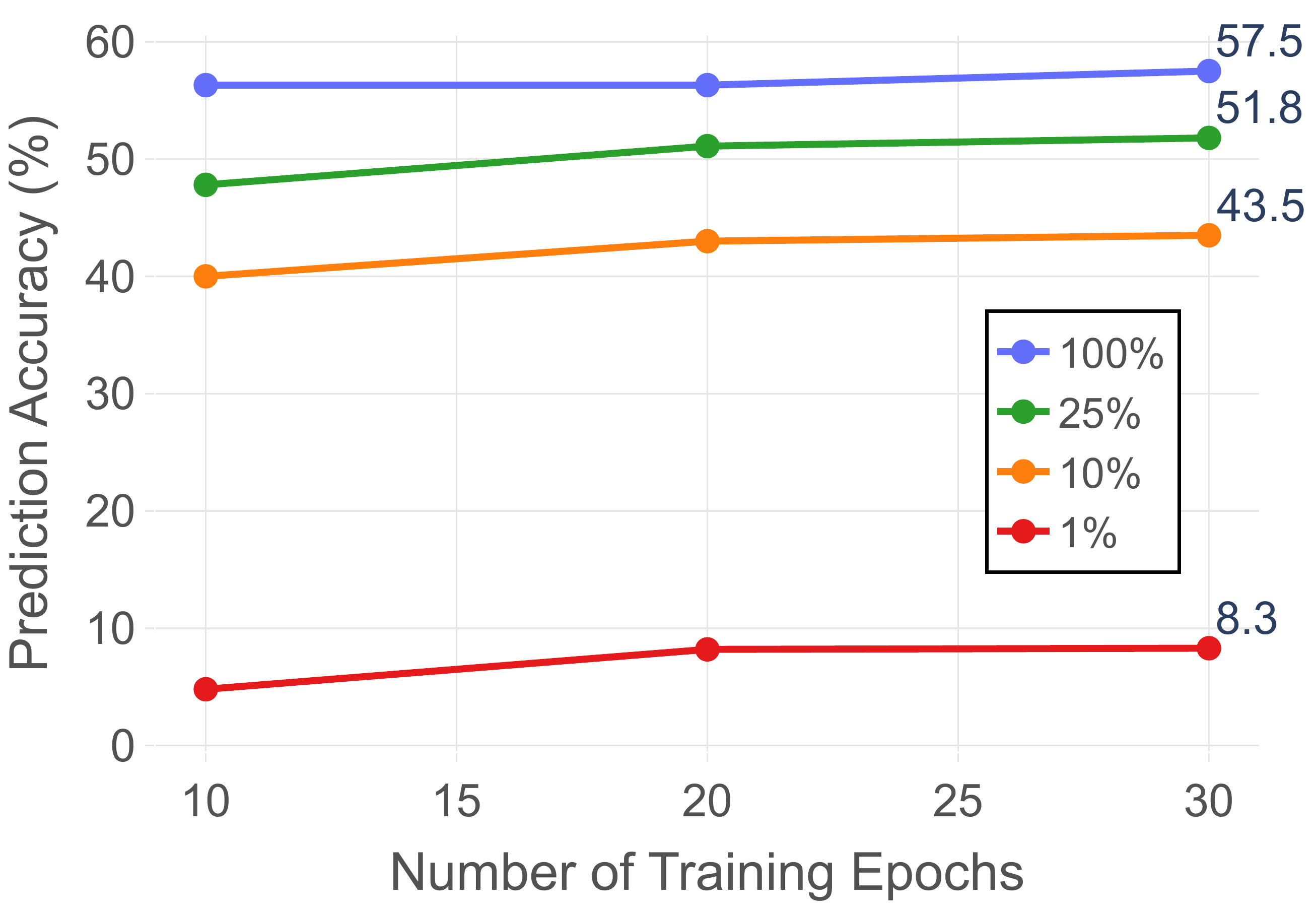}
\caption{\label{tab:data-dependence}\footnotesize Average prediction accuracy as a function of training set size (\textit{100\%, 25\%, 10\%, or 1\%} of the full training set) for the GPT-2 model on the test set.  Even with a large rediction in training data, the model is still able to accurrately predict a large number of visual semantic plans. Performance represents the permissive scoring metric in the \textit{``full minus first''} condition in Table~\ref{tab:performance}.
}
\end{figure}

\section{Data Dependence and Few-Shot Learning}
To examine how performance varies with the amount of training data available, we randomly downsampled the amount of training data to \textit{25\%, 10\%, and 1\%} of its original size.  This analysis, shown in Figure~\ref{tab:data-dependence}, demonstrates that relatively high performance on the visual semantic prediction task is still possible with comparatively little training data.  When only 10\% of the original training data is used, average prediction accuracy reduces by 24\%, but still reaches 44\%.  In the \textit{few-shot} case (1\% downsampling), where each of the 7 \textsc{ALFRED} tasks observes only 4 gold command sequences each (for a total of 12 natural language directives per task) during training, the GPT-2 model is still able to generate an accurate visual semantic plan in 8\% of cases.  Given that large pre-trained language models have been shown to encode a variety of commonsense knowledge \textit{as-is}, without fine-tuning \cite{petroni2019language}, it is possible that some of the model's few-shot performance on \textsc{ALFRED} may be due to an existing knowledge of similar common everyday tasks. 



\section{Conclusion}
We empirically demonstrate that detailed gold visual semantic plans can be generated for 26\% of unseen task directives in the \textsc{ALFRED} challenge using a large pre-trained language model \textit{without} visual input from the simulated environment, where 58\% can be generated if starting locations are known.  We envision these plans may be used either as-is, or as an initial ``hypothetical'' plan of how the model believes the task might be solved in a generic environment, that is then modified based on visual or other input from a specific environment to further increase overall accuracy. 

We release our planner code, data, predictions, and analyses for incorporation into end-to-end systems at: \url{http://github.com/cognitiveailab/alfred-gpt2/} .


%

\bibliographystyle{acl_natbib}
\bibliography{anthology,emnlp2020}

\end{document}